\pdfoutput=1

\documentclass[11pt]{article}

\usepackage[]{acl}

\usepackage{times}
\usepackage{latexsym}

\usepackage[T1]{fontenc}

\usepackage[utf8]{inputenc}

\usepackage{microtype}

\usepackage{inconsolata}

\usepackage{CJK}
\usepackage{graphicx}
\usepackage{tcolorbox}
\usepackage{fdsymbol}
\newcommand{\cname}{ProFTAP}
\newcommand{\ourllm}{Qwen-72B-Poet}

\title{Can AI Write Classical Chinese Poetry like Humans? An Empirical Study Inspired by Turing Test}

\author{Zekun Deng$^\clubsuit{}^\diamondsuit$, Hao Yang$^\heartsuit{}^\diamondsuit$ \and Jun Wang$^\clubsuit{}^\heartsuit{}^\diamondsuit$ \\
  $^\clubsuit$ Department of Information Management, Peking University \\
  $^\heartsuit$ Institute of Artificial Intelligence, Peking University \\
  $^\diamondsuit$ Research Center for Digital Humanities, Peking University \\
  \texttt{\{dzk,yanghao2008,junwang\}@pku.edu.cn}}

\begin{document}
\maketitle
\begin{abstract}
Some argue that the essence of humanity, such as creativity and sentiment, can never be mimicked by machines. This paper casts doubt on this belief by studying a vital question: Can AI compose poetry as well as humans? To answer the question, we propose \cname{}, a novel evaluation framework inspired by Turing test to assess AI's poetry writing capability. We apply it on current large language models (LLMs) and find that recent LLMs do indeed possess the ability to write classical Chinese poems nearly indistinguishable from those of humans. We also reveal that various open-source LLMs can outperform GPT-4 on this task. 
\end{abstract}

\section{Introduction}

Today's world sees a fierce discourse on the potentiality of artificial intelligence transcending that of humans \cite{chui2016machines,de2021ai}. While studies show that artificial intelligence (AI) has already outdo humans on certain tasks \cite{silver2017mastering,geminiteam2023}, many hold that the very essence of humanity, such as creativity and sentiment, can never be mimicked by machines \cite{MILLET2023107707}.

Among the areas which have not yet been dominated by AI is poetry. Poetry holds a profound significance in the realm of human art and civilization. 
Serving as a vessel of creativity, poetry encapsulates complex feelings and ideas in a condensed and evocative form, allowing for a deeper exploration of the human condition. 

In response to the recent debate, this paper focuses on a vital question: Can AI compose poetry as well as humans? However, this question is tricky to answer since there has hardly been an acknowledged definition of a good poem or a bad poem. In fact, from Aristotle's \emph{Poetics} to Confucius' sayings in \emph{Analects}, the way to judge a poem has remained a subject of dispute throughout the history. Thus, using existing methods to assess the good and bad of poetry certainly is not likely to convince everyone. It is clear that a more objective, sound and convincing alternative is needed, which can be both accepted by most people and conducted in practice. 

Therefore, in this paper, we propose a novel evaluation framework for evaluating AI's poetry writing capability. The framework is inspired by Turing test and relies on \emph{distinguishability} to measure the poetry composing ability of AI. By taking classical Chinese poetry as an exemplar, based on our experimental result, we argue that current large language models (LLMs) do indeed possess the ability to write poems \emph{nearly} indistinguishable from those of humans.

The main contributions of this paper are:
\begin{enumerate}
    \item We propose \cname{}, a novel framework for evaluating AI-generated poetry inspired by Turing test. \cname{} is more objective and rigorous while easier to implement than prior manual methods.
    \item We apply \cname{} to popular LLMs and reveal their capabilities of classical Chinese poetry generation.
    \item We show that finetuned open-source LLM can outperform GPT-4 on classical Chinese poetry generation and are able to write poems nearly indistinguishable from those authored by ancient Chinese poets. 
\end{enumerate}

\section{Related Works}

Prior works have utilized numerous methods for evaluating AI-generated poetry. Many studies adopt automatic approaches. \citet{ormazabal-etal-2022-poelm} used the perplexity of language model. \citet{liu-etal-2019-rhetorically} used BLEU to evaluate model fitness. They also use Rhetoric F1 score to measure the rhetorically accuracy and Distinct-1/2 \cite{li-etal-2016-diversity} to evaluate diversity. \citet{Deng_Wang_Liang_Chen_Xie_Zhuang_Wang_Xiao_2020} used the similarity of sentence-level embeddings of sentences within a poem to reflect coherency. 

More researches prefer human evaluation. \citet{zhang-lapata-2014-chinese} proposed to ask experts to rate the poems using a Likert 1–5 scale on four dimensions: fluency, coherence, meaning and poeticness. Numerous later works followed this approach \cite{ijcai2019p684,van-de-cruys-2020-automatic,Deng_Wang_Liang_Chen_Xie_Zhuang_Wang_Xiao_2020}. \citet{ijcai2018p0633} changed the poeticness dimension into aesthetics and added a new dimension called topic relevance, which is followed by \citet{liu-etal-2019-rhetorically}. \citet{bena-kalita-2019-introducing} used similar scale to assess creativity in machine-generated poems, but with different dimensions. \citet{ormazabal-etal-2022-poelm} ask human to rank poems written by computers, laymen and experts. \citet{10.1007/978-3-319-49685-6_4} adopted Feigenbaum Test to evaluate machine-generated poems. 


\section{\cname{}: An Evaluation Framework for AI-generated Poetry}

Whether a poem is good or not is a question without definitive answer. 
Despite numerous proposed criteria, few have yet proven entirely fulfilling. Therefore, instead of trying to seek for a universally recognized standard of good poem, we resort to a simple but much less controversial criterion: distinguishability. Rather than determining whether AI poems are as good as human's, we shift focus to investigate whether AIs can generate poems \emph{akin} to humans. Consequently, we can measure the poetry writing ability of AI to some extent without having to decide how meritorious a poem is.

Therefore, we propose a novel evaluation framework 
which we name as \textbf{Pro}babilistic \textbf{F}eigenbaum \textbf{T}est for \textbf{A}I-generated \textbf{P}oetry (\cname). 
Feigenbaum test \cite{feigenbaum2003some} is a variation of Turing test, which can be used where a computer tries to replicate a domain expert in a specific field. The core of our framework is based on Feigenbaum test. The procedures of \cname{} and its differences from previous methods are stated in the subsections below.

\subsection{Procedures}

The procedures for applying \cname{} to evaluate the poetry generation capability of AI models $\mathcal{M} = \{m_1, ..., m_M\}$ are as follows.

\textbf{Step 1: Obtaining titles as conditions.} AIs are not expected to freely generate poems of any topic, since this may lead to unfair comparison. 
Therefore, we obtain a set of poems authored by real humans and later force the AIs to generate poems on each one of and only on the titles that we provide. Formally, a poem $p_{i,j}$ consists of a title $t_j$ and the main text $c_j$ $(j=1,...T)$. The set of poems by real humans is denoted $\mathcal{P}^* = \{p_{*,1}, ..., p_{*,T}\}$. We collect the titles of the poems $\mathcal{T} = \{t_1, ..., t_T\}$.

\textbf{Step 2: Preparing AI models.} AIs need to be prepared to generate desired form of output. For LLMs, as an example, preparation may include designing appropriate prompts. 

\textbf{Step 3: Generation.} We use each model $m_i\in \mathcal{M}$ to generate the main text of a poem conditioned on title $t_j \in \mathcal{T}$. 
The resulting poem is denoted $p_{i,j}$.
The set of all generated poems are denoted $\mathcal{P} = \{p_{1,1},...,p_{M,T}\}$.

\textbf{Step 4: Post-processing and Anti-plagiarism.} Post-processing is conducted in case an AI does not fully comply with the requirements. For instance, the words explaining the poem are removed by programs with certain rules. The generated poems are also matched against a ancient poem database to find plagiarism. Any poems with duplication are re-generated until no overlap can be found. The poems after this step are still denoted $\mathcal{P}$.

\textbf{Step 5: Human Judgement.} A group of human judges 
are recruited. 
The judges are instructed to assign a probability (0.0 to 1.0) to each poem they are given which indicates how likely this poem is written by a real human, without referring to any other information and without the help of other people or computer. We mix real-human poems and AI-generated ones together, i.e. $\mathcal{U} = \mathcal{P^*} \cup \mathcal{P}$. We randomly shuffle $\mathcal{U}$ and distribute the poems randomly to human judges. Each poem is assigned to at least $K$ different human judges. The mean probability value by different judges of a poem $p_{i,j} (i=*,1,2,..,M)$ is denoted $q_{i,j}$.

\textbf{Step 6: Obtaining metric.} We derive the Receiver Operating Characteristic of human judges with respect to distinguishing real-human poems with poems by each AI. We calculate the area under the curve (AUC) corresponding to each model. The closer AUC is to 0.5, the more similar the AI's poems are to human's. 
Additionally, we apply Wilcoxon signed-rank test \cite{wilcoxon1992individual} on $q_{i,j}$ and $q_{*,j}$ for each model $m_i\in \mathcal{M}$ to evaluate whether the probability difference between human- and AI-authored poems conditioned on the same title is statistically significant. 

\subsection{Difference from Previous Evaluation Methods}
Our framework is more advantageous than previous methods in three aspects. Firstly, \cname{} is more objective and rigorous. Prior human evaluation usually require judgements of subjective concepts, such as poeticness, aesthetics and coherence, and different people interpret these concepts differently. Secondly, \cname{} requires relatively smaller human effort. Compared with scoring four to five dimensions to assess a single poem, our framework only needs one. Thirdly, \cname{} is much more reliable and interpretable than automatic approaches such as BLEU or perplexity. The involvement of human effort increases the soundness of evaluation.

Although \citet{10.1007/978-3-319-49685-6_4} have used Feigenbaum Test in evaluating machine poetry, our framework differs from it significantly in two ways. 
First, 
\cname{} is based on probability estimation of authorship rather than a yes/no answer, enabling more informative and detailed statistical analysis. Furthermore, our framework does not necessarily require humans to compose new poems and can just use existing ones in database. In the case of classical Chinese poetry, this brings extra benefit because 
classical poems composed nowadays are hardly on par with ancient ones.

\section{Experiments and Results}

\subsection{Human Judgements of Current LLMs}

We employ \cname{} to evaluate the classical Chinese poetry writing ability of major current LLMs. We recruited 13 human judges to conduct the evaluation. Most of them have higher education background relevant to classical Chinese poetics. We randomly chose 110 poems ($T=110$) from Souyun\footnote{\url{https://sou-yun.cn}}, the largest public database of classical Chinese poetry to date. Since the database contains approximately 1 million poems, it is impossible for any ordinary individual to memorize even a small portion of them. Thus, we can rule out the possibility that human judges might know by memory that a randomly sampled poem is authored by human. 

We selected 10 AI models for assessment. For proprietary LLMs, we chose GPT-3.5, GPT-4 \cite{openai2023gpt4} and Ernie-Bot-4.0\footnote{See more information on \url{https://cloud.baidu.com}.}, which are often considered the best-performing models on many tasks in Chinese language. For open-source LLMs, we chose Qwen-72B-Chat \cite{bai2023qwen}, Yi-34B-Chat\footnote{\url{https://huggingface.co/01-ai/Yi-34B-Chat}}, Qwen-14B-Chat \cite{bai2023qwen} and ChatGLM3-6B \cite{du-etal-2022-glm}, covering various scales. We also evaluate Xunzi-Qwen-Chat\footnote{\url{https://modelscope.cn/models/Xunzillm4cc/Xunzi-Qwen-Chat}}, an LLM continually pretrained on ancient Chinese corpora based on Qwen-7B. We use a same set of prompt and hyperparameters for all models. We also tested Jiuge \cite{zhipeng-etal-2019-jiuge}, a reputable AI system specialized in this task. See Appendix~\ref{sec:experiment} for more details on our experiments.

In addition to out-of-the-box LLMs, we also train a new LLM specialized in this task and include it in our evaluation. This model, which we name as \ourllm{}, is finetuned from Qwen-72B-Chat-Int4 using Souyun data. Refer to Appendix~\ref{sec:poetmodel} to see training details. 

\subsection{Results}

The AUC and result for Wilcoxon test is shown in Table~\ref{tab:wilcoxon}. 
Among all the models, \ourllm{} has the lowest AUC (0.541), which indicates that its poems are most similar to human's. Xunzi-7B-Chat is the second best LLM and beats Qwen-72B-Chat, which shows the effectiveness of continual pretraining. GPT-4 has a much lower AUC than GPT-3.5, and Ernie-Bot-4.0 underperforms GPT-4. Although Jiuge is not based on LLM, it still outperforms all prior LLMs we tested. Despite its larger scale, Yi-34B-Chat underperforms Qwen-14B-Chat.

The $p$ of \ourllm{} is 0.058, which means that there is no statistically significant difference between the poems by \ourllm{} and humans. In contrast, the $p$ of all other models are no greater than 3e-4, which are statistically significant. The results show that \ourllm{} can actually write classical Chinese poetry nearly indistinguishable from those of humans, at least to some extent.

\begin{table}
\centering
\begin{tabular}{lrcr}
\hline
\textbf{Model} & \textbf{\#Prm} & \textbf{AUC} & \textbf{W.T. $p$}\\
\hline
\ourllm{} & 72B & 0.541 & 0.058 \\
Jiuge & N/A & 0.632 & 3e-4 \\
Xunzi-7B-Chat & 7B & 0.633 & 8e-5 \\
Qwen-72B-Chat & 72B & 0.641 & 7e-5 \\ 
GPT-4 & N/A & 0.670 & 2e-6 \\ 
Ernie-Bot-4.0 & N/A & 0.732 & <1e-8 \\
Qwen-14B-Chat & 14B & 0.784 & <1e-8 \\ 
GPT-3.5 & N/A & 0.874 & <1e-8 \\ 
Yi-34B-Chat & 34B & 0.889 & <1e-8 \\ 
ChatGLM3-6B & 6B & 0.913 & <1e-8 \\ \hline
\end{tabular}
\caption{AUC and $p$-value of Wilcoxon test (W.T. $p$) of each model. \#Prm stands for number of parameters.}
\label{tab:wilcoxon}
\end{table}

\section{Discussion}

We further investigate how explicit features of AI-generated poems might impact how they are valued. We filter the generated poems according to 2 crucial factors: line length and character repetition. Since human-written poems \textit{usually}\footnote{None of the criteria are strict in classical Chinese poetry, and many highly-appreciated poems actually break these rules. However, people do often use these criteria for judgement.} have patterned line lengths while certain AI models often disregard the convention, it might play an important role in the evaluation of AI poetry. This is the exact same case for character repetition. We calculate the AUC of each model with poems that violate either criterion removed (see Table~\ref{tab:filter}). Clearly, removing AI poems with atypical line lengths generally makes them more similar to human poems, but character repetition does not matter as much.

\newcommand{\doa}{\textcolor[rgb]{0,0.8,0}{$\downarrow$}}
\newcommand{\upa}{\textcolor[rgb]{0.8,0,0}{$\uparrow$}}

\begin{table}
\centering
\begin{tabular}{lcc}
\hline
\textbf{Model} & \textbf{Lin. Len.} & \textbf{Cha. Rep.} \\
\hline
\ourllm{} & 0.525 \doa & 0.563 \upa \\
Jiuge & 0.646 \upa & 0.698 \upa \\
Xunzi-7B-Chat & 0.578 \doa & 0.599 \doa \\
Qwen-72B-Chat & 0.571 \doa & 0.610 \doa \\ 
GPT-4 & 0.623 \doa & - \\ 
Ernie-Bot-4.0 & 0.741 \upa & - \\
Qwen-14B-Chat & 0.717 \doa & - \\ 
GPT-3.5 & 0.809 \doa & - \\ 
Yi-34B-Chat & 0.699 \doa & 0.844 \doa \\ 
ChatGLM3-6B & 0.822 \doa & - \\ \hline
\end{tabular}
\caption{AUC after removing poems violating each criterion. "\upa" indicates higher AUC than original and "\doa" lower. "-" indicates sample size less than 10.}
\label{tab:filter}
\end{table}

\begin{figure}
\includegraphics[width=\columnwidth]{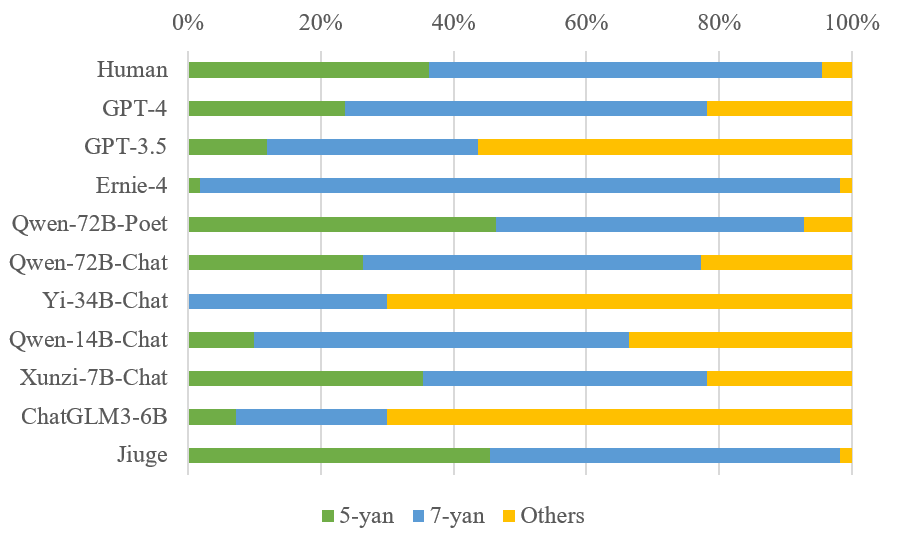}
\caption{Ratio of poems authored by each model and human in each category of yan.}
\label{fig:yan}
\end{figure}

\begin{table}
\centering
\begin{tabular}{lccc}
\hline
\textbf{Model} & \textbf{5-yan} & \textbf{7-yan} \\
\hline
\ourllm{} & 0.499 & 0.538 \\
Jiuge & 0.624 & 0.668 \\
Xunzi-7B-Chat & 0.520 & 0.601 \\
Qwen-72B-Chat & 0.578 & 0.571 \\ 
GPT-4 & 0.538 & 0.663 \\ 
Ernie-Bot-4.0 & - & 0.746 \\
Qwen-14B-Chat & 0.725 & 0.716 \\ 
GPT-3.5 & 0.851 & 0.777 \\ 
Yi-34B-Chat & - & 0.715 \\ 
ChatGLM3-6B & - & 0.764 \\ \hline
\end{tabular}
\caption{AUC of each category of line length. "-" indicates sample size less than 10.}
\label{tab:yan}
\end{table}

Since most classical Chinese poems have line length of 5 or 7 (i.e. 5-yan / 7-yan), we separate the poems into three categories: 5-yan, 7-yan and others. We calculate the ratio of generated poems in each category (Figure~\ref{fig:yan})\footnote{Jiuge requires yan to be designated before generation, so we randomly choose between 5 and 7 with same probability each time. However, the system occasionally returns poems neither 5-yan nor 7-yan even when it is designated to do so.} and the AUC of 5- and 7-yan poems with respect to human poems with the same line length (Table~\ref{tab:yan}). We find that there is not much difference between the AUC of 5- and 7-yan poems, but some models do have a tendency to generate more 7-yan over 5-yan.

It is also worth noting that, although our results show most of LLMs cannot write classical Chinese poems like human using the \emph{current} prompt, it is possible that these LLMs can do better if more advanced prompting techniques are used. We leave them for future work. 


\section{Conclusion}

We propose \cname{}, a new framework for evaluating AI-generated poetry, which is more objective and rigorous than previous methods. We apply \cname{} to popular LLMs and reveal that current AI models still have room for improvement in writing classical Chinese poems. However, we do find that finetuned open-source LLM can generate classical Chinese poems nearly indistinguishable from those of ancient Chinese poets. We expect our findings to be beneficial for future researches on AI poetry generation and evaluation.



\bibliography{anthology,custom}

\begin{thebibliography}{23}
\expandafter\ifx\csname natexlab\endcsname\relax\def\natexlab#1{#1}\fi

\bibitem[{Bai et~al.(2023)Bai, Bai, Chu, Cui, Dang, Deng, Fan, Ge, Han, Huang et~al.}]{bai2023qwen}
Jinze Bai, Shuai Bai, Yunfei Chu, Zeyu Cui, Kai Dang, Xiaodong Deng, Yang Fan, Wenbin Ge, Yu~Han, Fei Huang, et~al. 2023.
\newblock Qwen technical report.
\newblock \emph{arXiv preprint arXiv:2309.16609}.

\bibitem[{Bena and Kalita(2019)}]{bena-kalita-2019-introducing}
Brendan Bena and Jugal Kalita. 2019.
\newblock \href {https://aclanthology.org/2019.icon-1.4} {Introducing aspects of creativity in automatic poetry generation}.
\newblock In \emph{Proceedings of the 16th International Conference on Natural Language Processing}, pages 26--35, International Institute of Information Technology, Hyderabad, India. NLP Association of India.

\bibitem[{Chen et~al.(2019)Chen, Yi, Sun, Li, Yang, and Guo}]{ijcai2019p684}
Huimin Chen, Xiaoyuan Yi, Maosong Sun, Wenhao Li, Cheng Yang, and Zhipeng Guo. 2019.
\newblock \href {https://doi.org/10.24963/ijcai.2019/684} {Sentiment-controllable chinese poetry generation}.
\newblock In \emph{Proceedings of the Twenty-Eighth International Joint Conference on Artificial Intelligence, {IJCAI-19}}, pages 4925--4931. International Joint Conferences on Artificial Intelligence Organization.

\bibitem[{Chui et~al.(2016)Chui, Manyika, and Miremadi}]{chui2016machines}
Michael Chui, James Manyika, and Mehdi Miremadi. 2016.
\newblock Where machines could replace humans-and where they can't (yet).
\newblock \emph{The McKinsey Quarterly}, pages 1--12.

\bibitem[{De~Cremer and Kasparov(2021)}]{de2021ai}
David De~Cremer and Garry Kasparov. 2021.
\newblock Ai should augment human intelligence, not replace it.
\newblock \emph{Harvard Business Review}, 18:1.

\bibitem[{Deng et~al.(2020)Deng, Wang, Liang, Chen, Xie, Zhuang, Wang, and Xiao}]{Deng_Wang_Liang_Chen_Xie_Zhuang_Wang_Xiao_2020}
Liming Deng, Jie Wang, Hangming Liang, Hui Chen, Zhiqiang Xie, Bojin Zhuang, Shaojun Wang, and Jing Xiao. 2020.
\newblock \href {https://doi.org/10.1609/aaai.v34i05.6265} {An iterative polishing framework based on quality aware masked language model for chinese poetry generation}.
\newblock \emph{Proceedings of the AAAI Conference on Artificial Intelligence}, 34(05):7643--7650.

\bibitem[{Dettmers et~al.(2023)Dettmers, Pagnoni, Holtzman, and Zettlemoyer}]{dettmers2023qlora}
Tim Dettmers, Artidoro Pagnoni, Ari Holtzman, and Luke Zettlemoyer. 2023.
\newblock Qlora: Efficient finetuning of quantized llms.
\newblock \emph{arXiv preprint arXiv:2305.14314}.

\bibitem[{Du et~al.(2022)Du, Qian, Liu, Ding, Qiu, Yang, and Tang}]{du-etal-2022-glm}
Zhengxiao Du, Yujie Qian, Xiao Liu, Ming Ding, Jiezhong Qiu, Zhilin Yang, and Jie Tang. 2022.
\newblock \href {https://doi.org/10.18653/v1/2022.acl-long.26} {{GLM}: General language model pretraining with autoregressive blank infilling}.
\newblock In \emph{Proceedings of the 60th Annual Meeting of the Association for Computational Linguistics (Volume 1: Long Papers)}, pages 320--335, Dublin, Ireland. Association for Computational Linguistics.

\bibitem[{Feigenbaum(2003)}]{feigenbaum2003some}
Edward~A Feigenbaum. 2003.
\newblock Some challenges and grand challenges for computational intelligence.
\newblock \emph{Journal of the ACM (JACM)}, 50(1):32--40.

\bibitem[{Google(2023)}]{geminiteam2023}
Gemini~Team Google. 2023.
\newblock \href {http://arxiv.org/abs/2312.11805} {Gemini: A family of highly capable multimodal models}.
\newblock \emph{arXiv preprint arXiv:2312.11805}, arXiv:2312.11805.

\bibitem[{Li et~al.(2016)Li, Galley, Brockett, Gao, and Dolan}]{li-etal-2016-diversity}
Jiwei Li, Michel Galley, Chris Brockett, Jianfeng Gao, and Bill Dolan. 2016.
\newblock \href {https://doi.org/10.18653/v1/N16-1014} {A diversity-promoting objective function for neural conversation models}.
\newblock In \emph{Proceedings of the 2016 Conference of the North {A}merican Chapter of the Association for Computational Linguistics: Human Language Technologies}, pages 110--119, San Diego, California. Association for Computational Linguistics.

\bibitem[{Liu et~al.(2019)Liu, Fu, Cao, de~Melo, Tam, Niu, and Zhou}]{liu-etal-2019-rhetorically}
Zhiqiang Liu, Zuohui Fu, Jie Cao, Gerard de~Melo, Yik-Cheung Tam, Cheng Niu, and Jie Zhou. 2019.
\newblock \href {https://doi.org/10.18653/v1/P19-1192} {Rhetorically controlled encoder-decoder for {M}odern {C}hinese poetry generation}.
\newblock In \emph{Proceedings of the 57th Annual Meeting of the Association for Computational Linguistics}, pages 1992--2001, Florence, Italy. Association for Computational Linguistics.

\bibitem[{Millet et~al.(2023)Millet, Buehler, Du, and Kokkoris}]{MILLET2023107707}
Kobe Millet, Florian Buehler, Guanzhong Du, and Michail~D. Kokkoris. 2023.
\newblock \href {https://doi.org/https://doi.org/10.1016/j.chb.2023.107707} {Defending humankind: Anthropocentric bias in the appreciation of ai art}.
\newblock \emph{Computers in Human Behavior}, 143:107707.

\bibitem[{OpenAI(2023)}]{openai2023gpt4}
OpenAI. 2023.
\newblock \href {http://arxiv.org/abs/2303.08774} {Gpt-4 technical report}.

\bibitem[{Ormazabal et~al.(2022)Ormazabal, Artetxe, Agirrezabal, Soroa, and Agirre}]{ormazabal-etal-2022-poelm}
Aitor Ormazabal, Mikel Artetxe, Manex Agirrezabal, Aitor Soroa, and Eneko Agirre. 2022.
\newblock \href {https://doi.org/10.18653/v1/2022.findings-emnlp.268} {{P}oe{LM}: A meter- and rhyme-controllable language model for unsupervised poetry generation}.
\newblock In \emph{Findings of the Association for Computational Linguistics: EMNLP 2022}, pages 3655--3670, Abu Dhabi, United Arab Emirates. Association for Computational Linguistics.

\bibitem[{Silver et~al.(2017)Silver, Schrittwieser, Simonyan, Antonoglou, Huang, Guez, Hubert, Baker, Lai, Bolton et~al.}]{silver2017mastering}
David Silver, Julian Schrittwieser, Karen Simonyan, Ioannis Antonoglou, Aja Huang, Arthur Guez, Thomas Hubert, Lucas Baker, Matthew Lai, Adrian Bolton, et~al. 2017.
\newblock Mastering the game of go without human knowledge.
\newblock \emph{Nature}, 550(7676):354--359.

\bibitem[{Touvron et~al.(2023)Touvron, Martin, Stone, Albert, Almahairi, Babaei, Bashlykov, Batra, Bhargava, Bhosale et~al.}]{touvron2023llama}
Hugo Touvron, Louis Martin, Kevin Stone, Peter Albert, Amjad Almahairi, Yasmine Babaei, Nikolay Bashlykov, Soumya Batra, Prajjwal Bhargava, Shruti Bhosale, et~al. 2023.
\newblock Llama 2: Open foundation and fine-tuned chat models.
\newblock \emph{arXiv preprint arXiv:2307.09288}.

\bibitem[{Van~de Cruys(2020)}]{van-de-cruys-2020-automatic}
Tim Van~de Cruys. 2020.
\newblock \href {https://doi.org/10.18653/v1/2020.acl-main.223} {Automatic poetry generation from prosaic text}.
\newblock In \emph{Proceedings of the 58th Annual Meeting of the Association for Computational Linguistics}, pages 2471--2480, Online. Association for Computational Linguistics.

\bibitem[{Wang et~al.(2016)Wang, Luo, and Wang}]{10.1007/978-3-319-49685-6_4}
Qixin Wang, Tianyi Luo, and Dong Wang. 2016.
\newblock Can machine generate traditional chinese poetry? a feigenbaum test.
\newblock In \emph{Advances in Brain Inspired Cognitive Systems}, pages 34--46, Cham. Springer International Publishing.

\bibitem[{Wilcoxon(1992)}]{wilcoxon1992individual}
Frank Wilcoxon. 1992.
\newblock Individual comparisons by ranking methods.
\newblock In \emph{Breakthroughs in Statistics: Methodology and Distribution}, pages 196--202. Springer.

\bibitem[{Yi et~al.(2018)Yi, Sun, Li, and Yang}]{ijcai2018p0633}
Xiaoyuan Yi, Maosong Sun, Ruoyu Li, and Zonghan Yang. 2018.
\newblock \href {https://doi.org/10.24963/ijcai.2018/633} {Chinese poetry generation with a working memory model}.
\newblock In \emph{Proceedings of the Twenty-Seventh International Joint Conference on Artificial Intelligence, {IJCAI-18}}, pages 4553--4559. International Joint Conferences on Artificial Intelligence Organization.

\bibitem[{Zhang and Lapata(2014)}]{zhang-lapata-2014-chinese}
Xingxing Zhang and Mirella Lapata. 2014.
\newblock \href {https://doi.org/10.3115/v1/D14-1074} {{C}hinese poetry generation with recurrent neural networks}.
\newblock In \emph{Proceedings of the 2014 Conference on Empirical Methods in Natural Language Processing ({EMNLP})}, pages 670--680, Doha, Qatar. Association for Computational Linguistics.

\bibitem[{Zhipeng et~al.(2019)Zhipeng, Yi, Sun, Li, Yang, Liang, Chen, Zhang, and Li}]{zhipeng-etal-2019-jiuge}
Guo Zhipeng, Xiaoyuan Yi, Maosong Sun, Wenhao Li, Cheng Yang, Jiannan Liang, Huimin Chen, Yuhui Zhang, and Ruoyu Li. 2019.
\newblock \href {https://doi.org/10.18653/v1/P19-3005} {{J}iuge: A human-machine collaborative {C}hinese classical poetry generation system}.
\newblock In \emph{Proceedings of the 57th Annual Meeting of the Association for Computational Linguistics: System Demonstrations}, pages 25--30, Florence, Italy. Association for Computational Linguistics.

\end{thebibliography}

\appendix

\section{Experimental Details}
\label{sec:experiment}
\subsection{Prompts and Hyperparameters}
\label{sec:prompt}

For all LLMs we assessed (except \ourllm{}), we use the same prompt as below:

\begin{tcolorbox}[width=\linewidth,
                  boxsep=0pt,
                  left=5pt,
                  right=5pt,
                  top=5pt,
                  colback=white,
                  boxrule=1pt,
                  ]
\begin{CJK*}{UTF8}{gkai}想象你是一位著名诗人，请你写一首题为《\verb|{{title}}|》的古诗。要让别人以为你的诗是真人所作，不要让人看出是机器生成的。\end{CJK*}

(English translation: \emph{Imagine you are a famous poet. Please write a classical poem titled} \verb|{{title}}|. \emph{Lead people to believe that your poem is written by a real human and do not let them realize it is generated by a machine.})
\end{tcolorbox}

We attempted to use this prompt with LLaMA 2 \cite{touvron2023llama} chat models of all different sizes, but they refuse to respond due to their safety restrictions, so we are not able to test them.

We use a temperature of 0.9 during generation. We keep other hyperparameters like top-p to the default of each model. We do not apply any other modifications.

\subsection{Usage of Commercial APIs}

We use API to access proprietary LLMs. The exact model version of GPT-4 we used is \verb|gpt-4-1106-preview| (also known as GPT-4 Turbo). The version of GPT-3.5 is \verb|gpt-3.5-turbo-1106|. We accessed Ernie-Bot-4.0 API in November 2023.

\subsection{Experimental Process}

We use Souyun database for anti-plagiarism. We consider a poem duplicative if there are no less than two consecutive lines that match exactly with line from the database. We convert all text into Simplified Chinese. The minimum number of human judges by whom each poem is assessed is 2 ($K=2$). 

\section{Training of \ourllm{}}
\label{sec:poetmodel}

We filtered poems from Souyun according to 2 criteria: 
\begin{enumerate}
    \item The poem is written in Tang (\begin{CJK*}{UTF8}{gbsn}唐\end{CJK*}) dynasty.
    \item The form of the poem is either \begin{CJK*}{UTF8}{gbsn}\emph{gufeng}(古风), \emph{lushi}(律诗), \emph{jueju}(绝句), \emph{siyan}(四言), \emph{liuyan}(六言) or \emph{pailu}(排律)\end{CJK*}.
\end{enumerate}
We sample 96\% of them as training data, yielding 40,026 training samples.

We convert poems to ChatML format as follows.
\begin{tcolorbox}[width=\linewidth,
                  boxsep=0pt,
                  left=5pt,
                  right=5pt,
                  top=5pt,
                  colback=white,
                  boxrule=1pt,
                  ]
\verb=<|im_start|>=system\verb|\n| \\
You are a helpful assistant.\verb=<|im_end|>=\verb|\n| \\
\verb=<|im_start|>=user\verb|\n| \\
\begin{CJK*}{UTF8}{gkai}写一首题为《\verb|{{title}}|》的\verb|{{form}}|。\end{CJK*} \verb=<|im_end|>=\verb|\n| \\
\verb=<|im_start|>=assistant\verb|\n| \\
\verb|{{content}}|\verb=<|im_end|>=\verb|\n| 
\end{tcolorbox}
Here, \verb|{{form}}| is the form of poem, which can be \begin{CJK*}{UTF8}{gbsn}\emph{gufeng}(古风), \emph{lushi}(律诗) or \emph{jueju}(绝句)\end{CJK*}. We finetune from \verb|Qwen-72B-Chat-Int4| model. We adopt 4-bit QLoRA \cite{dettmers2023qlora} to improve efficiency. The QLoRA rank is 64 and alpha is 16. We set batch size to 1 and gradient accumulation steps to 8. We set max sequence length to 512. The learning rate is 1e-4 with 256 warm-up steps. We train the model for 1.5 epochs. The training took 70 hours on one Nvidia A100 GPU.

The prompt during evaluation is identical to the one used in training. During evaluation, we randomly choose one form out of the three with equal probability for each prompt. 

Notably, we find that when the model is instructed to write a poem on a title and form that are part of the training data, it tends to recite the original poem in the training set. We intend to address this issue in the future.

\newcommand{\songti}[1]{\begin{CJK*}{UTF8}{gbsn}#1\end{CJK*}}
\newcommand{\kaiti}[1]{\begin{CJK*}{UTF8}{gkai}#1\end{CJK*}}

\section{Generation Examples}
\label{sec:example}

Below are 2 poems generated by \ourllm{}.

\noindent \textbf{(1) \songti{夜雪} (Night Snow)} \\
\kaiti{已觉衾枕冷，转见窗户明。积雪阴云尽，寒飙曙色清。林疏犹现月，岩深可听声。唯应御楼望，遥辽近禁城。} \\
\emph{Already feeling the bed and pillow cold, then seeing the bright light outside the window. The accumulated snow disperses the dark clouds, and the cold wind gradually makes the dawn clear. The sparse trees still reveal the moonlight, and faint sounds come from deep within the mountain rocks. Only by gazing from the imperial tower can one approach the distant Forbidden City.}\\
\textbf{Analysis: } The poem adopts a white sketch technique to depict the snowy scenery, incorporating the "cold," "brightness," and "stillness" of the night snow, creating a distinctive style. Although the poem describes the scenery, it focuses on the word "night," portraying the sights and feelings of a snowy night, blending emotion into the scenery.

The opening two lines borrow from Bai Juyi(\songti{白居易})'s phrases "\kaiti{已讶衾枕冷，复见窗户明}", yet refine the wording with innovative expression.

In the opening line, "Already feeling the bed and pillow cold," the poet describes the initial fall of night snow, where the poet has already sensed the chill in the bed and pillow, indicating that the snow has been falling for quite some time. The poem uses the perspective of human sensations, making the "night snow" tangible.

Continuing with "seeing the bright light," the poet looks out from the window, noticing that the snowfall shows no signs of diminishing, causing the poet to feel that the window has become brighter, while moonlight and lamplight cannot achieve the same effect. The use of the word "\kaiti{转}" (turn) is clever, revealing the entire process of the poet's perception of the snowy scenery from deep night to daybreak, emphasizing the poet's sensitivity to the snow.

Subsequently, the poet sees through the window, "The accumulated snow disperses the dark clouds, and the cold wind gradually makes the dawn clear." The snowfall is significant, lasting throughout the night. By daybreak, the snow stops, the sun emerges, the accumulated snow and dark clouds disappear, and the cold northern wind subsides, leaving a clear dawn illuminating the earth.

"Trees sparse, still showing the moon; rocks deep, able to hear sounds." Stepping outside, the poet observes sparse trees with the moon still shining, and deep rocks where one can hear the sound of the wind. The poet not only portrays the night scene with clear moonlight but also conveys the serene night with audible stillness in the deep rocks.

In the final line, "Only from the imperial tower can one see, distant and close to the Forbidden City," it indicates that one can only appreciate the distant snowy scenery near the Forbidden City by ascending the imperial tower. 

The poem depicts both distant and near views, still and dynamic scenes, brightness and darkness, snow and moon scenery, night and dawn scenes. It portrays the night snow not directly but indirectly, focusing on human sensations with "bed and pillow cold" and "window bright," providing a vivid impression. In terms of structure, the poet skillfully uses the technique of side description, with "The accumulated snow disperses the dark clouds, and the cold wind gradually makes the dawn clear" corresponding to "Already feeling the bed and pillow cold, seeing the window bright," creating a cohesive and well-structured poem. Additionally, the poem employs metaphor and contrast to make the night snow more tangible, contrasting "The cold wind gradually makes the dawn clear" with "The accumulated snow disperses the dark clouds."

Therefore, this poem demonstrates a skillful technique, a well-organized logic, and a clear and understandable content. It reads as if it were crafted by a real person, making it difficult to discern that it was written by a machine solely based on its content.

\noindent \textbf{(2) \songti{游三清山} (Roaming the Three Pure Peaks)} \\
\kaiti{昔闻三清山，今始到其境。石壁屹当路，万古雪霜冷。山灵有神功，斧凿不可整。巨灵擘两掌，裂破混沌景。我欲呼神工，移置华山顶。} \\
\emph{Long had I heard of the Three Pure Peaks, today I stand in their midst. Cliffs rise imposingly along the path, bearing witness to millennia of frost and cold. The spirits of this mountain wield mystical forces, impervious to the chiseling of axes. The giant spirit stretches forth its palms, tearing apart the chaotic scenes. I wish to summon that miraculous artisan, to relocate the Three Pure Peaks to the summit of Mount Hua.}\\
\textbf{Analysis: } The poem depicts the magnificent scenery of the Three Pure Peaks, expressing the author's sincere admiration for nature, in harmony with the Taoist reverence for the natural order symbolized by the Three Pure Peaks. The language of the entire poem is simple, the imagery is grand, and the style is elevated.

"Long had I heard of the Three Pure Peaks, today I stand in their midst." The poem begins with a discourse, and the phrase "Long had I heard" sets a lofty and powerful tone for the entire poem. The three words "Three Pure Peaks" not only clarify the title but also lead into the subsequent verses. The phrase "today I stand in their midst" informs us of the time the poet visited the Three Pure Peaks, indicating a transition from hearing about them to actually experiencing them. The word "today" suggests the poet's journey was not an easy one, laying the groundwork for the subsequent admiration of the spectacular scenery.

The poet then proceeds to describe the scenery of the Three Pure Peaks in two poetic lines. "Cliffs rise imposingly along the path, bearing witness to millennia of frost and cold." The most striking aspect of the Three Clear Peaks is their peculiar peaks and rocks, and the exaggerated phrase "millennia of frost and cold" vividly conveys the perennial snow and the naturally severe environment. These two lines succinctly and vividly depict the scenery, striking a balance between conciseness and vividness.

Using an exaggerated style, the author then portrays the extraordinary features of the Three Pure Peaks. "The mountain spirits possess mystical powers, impervious to the chiseling of axes." This vividly conveys that the Three Clear Peaks are the result of supernatural craftsmanship, steep and rugged without any human intervention, naturally formed. "The giant spirit stretches forth its palms, tearing apart the chaotic scenes," refers to the mythical giant spirit who split open the scenes of chaos on Mount Hua. The poem, by drawing on mythology, further emphasizes the ruggedness and towering nature of the Three Pure Peaks.

The final two lines, "I wish to summon that miraculous artisan, to relocate the Three Pure Peaks to the summit of Mount Hua," echo the opening lines. The poet, witnessing the divine craftsmanship of the Three Clear Peaks, desires to call upon the artisan to move them to the summit of Mount Hua, enhancing the beauty of Mount Hua and expressing the author's deep affection for the Three Pure Peaks.

The poem combines reality and imagination, with the first half describing the scenery of the Three Pure Peaks and the second half portraying the poet's imaginative scene. The poem seamlessly transitions between the two, creating a cohesive whole. Beginning with a discourse, then depicting the peculiar peaks and rocks, and finally weaving in mythical legends, the poem paints a picture of the Three Pure Peaks as lofty, perilous, with peculiar peaks and rocks, evoking a sense of awe and fascination.

This poem bears the title of "Three Pure Peaks," yet it does not focus on describing the scenery of the Three Pure Peaks. Instead, it approaches the subject from the perspective of the immortals symbolized by the Three Pure Ones, explaining that the scenery of the Three Pure Peaks is naturally formed, not the result of human intervention. Although it appears to be a travel poem, it completely avoids cliches, giving it an authentic quality that is difficult to distinguish in terms of technique-whether it was written by a machine or a human.

\end{document}